\documentclass[12pt, journal, onecolumn]{IEEEtran}
\usepackage{latexsym,epsfig,color,graphics,makeidx,amsfonts,subfigure}

\newcommand{\rA}{\ensuremath{\rightarrow}}
\newcommand{\sm}{\mbox{\textendash}}

\date{}

\begin{document}

\title{Temporal data mining for root-cause analysis of machine faults in
automotive assembly lines}
\author{Srivatsan~Laxman, Basel~Shadid, P.~S.~Sastry and K.~P.~Unnikrishnan
\thanks{S.~Laxman is with Microsoft Research, Bangalore. This work was done when he was with
Indian Institute of Science, Bangalore.}
\thanks{B.~Shadid is with General Motors, St. Catherines.}
\thanks{P.~S.~Sastry is with Indian Institute of Science, Bangalore.}
\thanks{K.~P.~Unnikrishnan is with Wayne State University, Detroit. This work was done when
he was with General Motors R\&D center, Warren.}}
%\institute{Microsoft Research, Bangalore \and Indian Institute of Science, Bangalore, India \and General Motors, St. Catherines, Canada, \and  General Motors, Warren, USA}

\maketitle

\begin{abstract}
Engine assembly is a complex and heavily automated distributed-control process, with large amounts of faults data logged everyday. We describe an application of temporal data mining for analyzing fault logs in an engine assembly plant. Frequent episode discovery framework is a model-free method that can be used to deduce (temporal) correlations among events from the logs in an efficient manner. In addition to being theoretically elegant and computationally efficient, frequent episodes are also easy to interpret in the form actionable recommendations. Incorporation of domain-specific information is critical to successful application of the method for analyzing fault logs in the manufacturing domain. We show how domain-specific knowledge can be incorporated using heuristic rules that act as pre-filters and post-filters to frequent episode discovery. The system described here is currently being used in one of the engine assembly plants of General Motors and is planned for adaptation in other plants. To the best of our knowledge, this paper presents the first real, large-scale application of temporal data mining in the manufacturing domain. We believe that the ideas presented in this paper can help practitioners engineer tools for analysis in other similar or related application domains as well.

\end{abstract}

\section{Introduction}

Automotive engine assembly is a heavily automated and complex process that is 
controlled in a distributed fashion. 
 Each assembly plant consists of several machines (or
operations/stations) that are extensively inter-connected and are programmed to
automatically execute the various operations necessary to manufacture an automotive
engine. The distributed control system maintains elaborate logs regarding
the time-evolving conditions of all machines in the plant, the status of
different operations performed on a particular engine, the throughput statistics
of the plant, etc. In this paper, we present an application that uses
temporal data mining techniques for analyzing these time-stamped logs to help in 
fault analysis and root-cause diagnosis. The application presented here is
currently being used on a regular basis in engine assembly plants of
General Motors.

%\begin{figure}[t]
%\centering
%\subfigure[Layout of a typical engine assembly line.]{
%\includegraphics[width=4in]{lomo.eps}
%%\caption{Layout of a typical engine assembly line.}
%\label{fig:lomo}
%}
%\subfigure[Snap-shot of data]{
%\includegraphics[height=3in]{datasnapshot.eps}
%\label{fig:datasnapshot}
%}
%\end{figure}

\begin{figure}[t]
\centering
\includegraphics[width=6in]{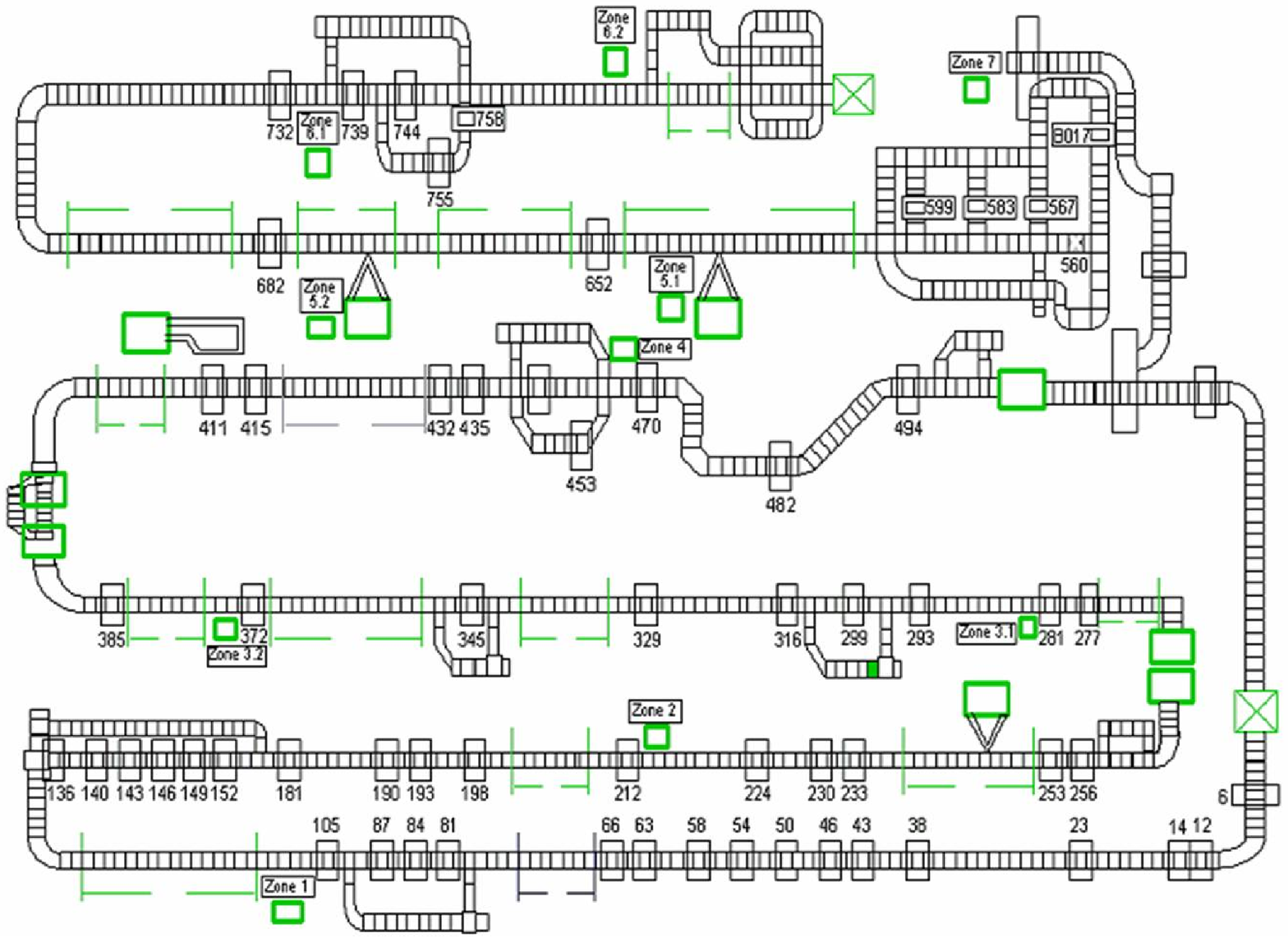}
\caption{Layout of a typical engine assembly line: A line is divided into several zones, indicated
above as, {\em Zone~1}, {\em Zone~2}, {\em Zone~3.1}, etc. Each zone consists of several machines or
stations, indicated in the plan through unique numbers. At a given time, typically, multiple stations
in the same (or different) zone(s) may be engaged, simultaneously operating on several
engine blocks at various stages of completion.}
\label{fig:lomo}
\end{figure}

\subsection{Engine plant data}
\label{sec:eapd}

The data records in manufacturing systems are mainly time-stamped 
records that take one of two forms: event-based records; time-based 
records. Machine faults logs are event-based records that report change 
in machine state from running state to down state. These records signify 
the behaviour observations of a machine as it unfolds, but lack the 
resolution that is needed to represent the dynamics of the change.

An engine plant consists of several machining and assembly lines. Each
line is divided into several zones which are usually separated by off-line
buffers. A zone contains a group of machines which are physically
and/or logically interrelated. Fig.~\ref{fig:lomo} shows the layout of a
typical engine assembly line. Engine manufacturing is a sequential process
as the engine and its components undergo a sequence of operations. Each
machine performs one or more operations on the engine. The line is
controlled by a  distributed control system.  Each
operation itself is divided into a sequence of controlled steps. For
example, the `drill operation' can be divided into the following steps:
clamp the part, advance the tool, start drilling, monitor final depth,
return the tool and unclamp the part. All these are controlled steps
in the sense that there are some conditions prescribed for
each step which must be satisfied to move to the following step; 
otherwise a machine fault occurs. For
example, associated with every step is a time limit. If the time taken
by a step exceeds this limit, a machine fault occurs.  Another reason
why a machine fault may be reported is if some precondition of the
operation step changes during the operation.  For example, the part
can get unclamped while the tool is still engaged. All such fault
conditions identified automatically by the control system are logged
into appropriate databases.  In general, faults would result in reduced
throughput (because, e.g., some operations may have to be redone) or may
even result in stoppage of the line. The plant floor engineers have to
constantly monitor the fault alarms and decide on the manual corrective
actions needed to keep the line running smoothly.

\begin{figure}[t]
\centering
\includegraphics[height=4in]{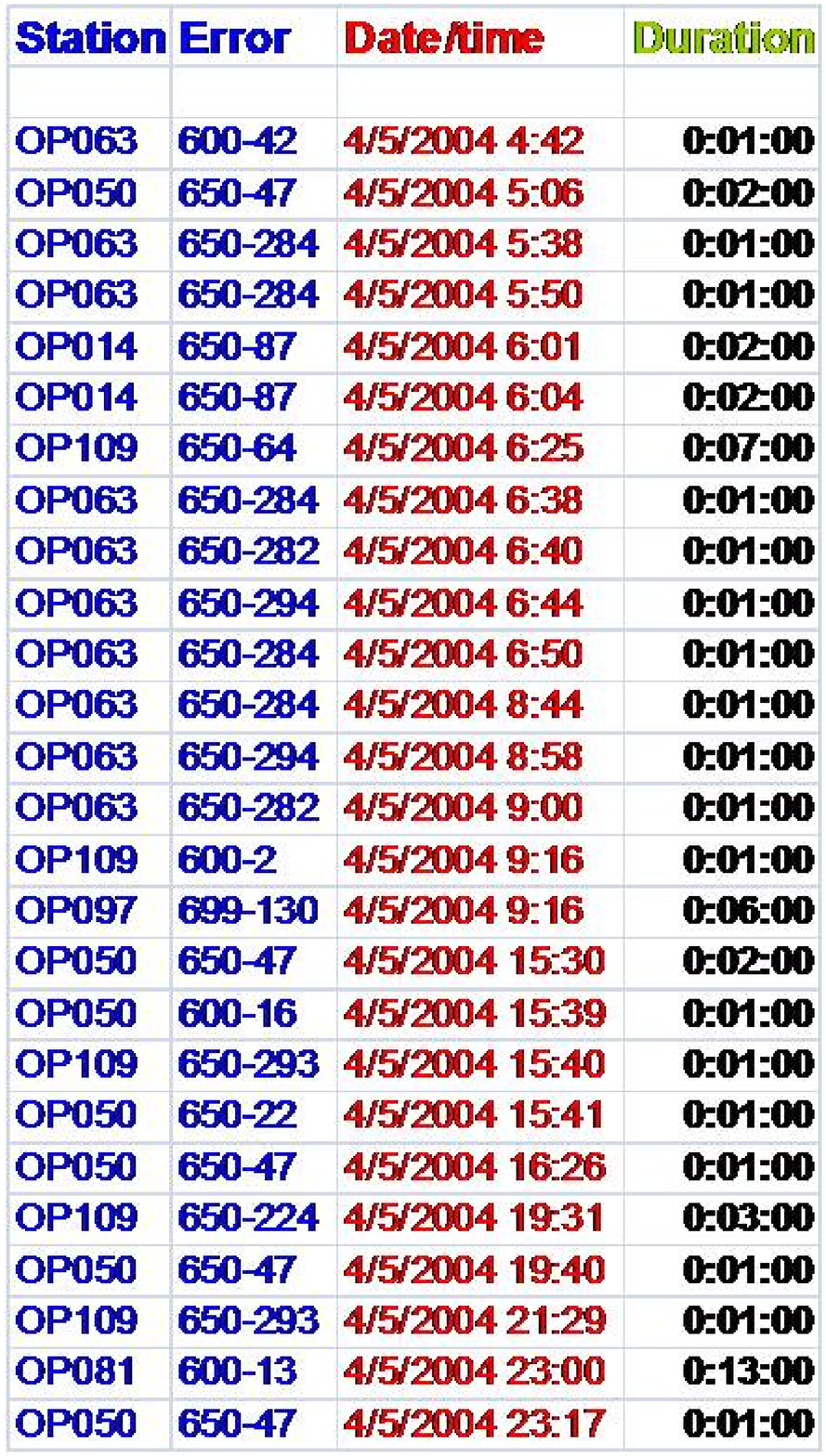}
\caption{A snapshot of the engine plant data logs. Each record has four fields: station name, error code, start
time of fault and duration of fault. The error code is obtained by combining subsystem code with the
exact fault code.}
\label{fig:datasnapshot}
\end{figure}

The machine fault logs are a time-ordered sequence of faults that have
occurred.  Each station or machine is identified by a {\em machine} or
{\em station code}. Whenever possible, the exact {\em subsystem} in
the machine that reports the fault is also recorded. For each different
fault that a machine can report there is a unique {\em fault} (or {\em error code}). Thus,
each fault record in the log has the following fields: (1)~operation,
(2)~subsystem, (3)~fault, (4)~occurred time, and (5)~resolved time. 
The first three fields take values from a finite, hierarchically arranged
alphabet and the last two time fields record the corresponding date and
time up to a resolution of one second. Fig.~\ref{fig:datasnapshot}
shows a snapshot of the data logs. Here the subsystem and 
fault codes are combined under the
``error'' column. Also, this snapshot quotes ``durations'' of the faults 
(computed as the difference between
occurred and resolved times) instead of the start and end times.
The control system is programmed to recover from some faults automatically 
while for some other faults manual intervention is necessary. The fault logs 
record both types of faults. 
%Also, when manual interventiona is necessary, 
%the operator can only see the last fault that brought the system to 
%an unrecoverable state; this fault may or may not be the root cause 
%for the problem. 

\begin{figure}[t]
\centering
\includegraphics[width=6in]{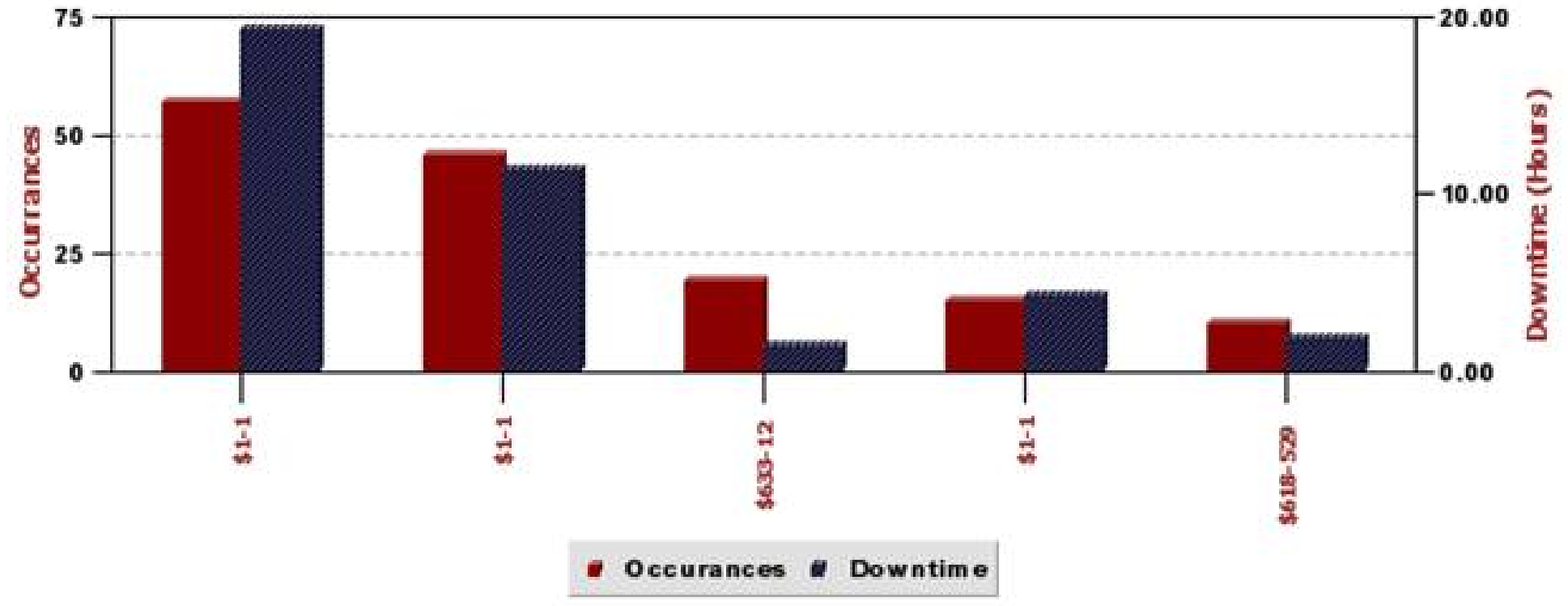}
\caption{An example of static analysis of machine fault logs that is used in the engine assembly
lines. The graph depicts the number of occurrences and the downtime (in hours) for different fault
conditions. Different fault conditions are marked on the X-axis. Number of occurrences and downtime
are depicted using red and blue-colored bars respectively.}
\label{fig:single-variable}
\end{figure}

\subsection{Static analysis of machine fault logs}

In the past,  machine fault
logs were used in the work floor in many ways. The simplest way was to rank
faults that occurred
in a line based on their frequencies in some given period (say, for
example, in the immediate past week). Another way was to rank them in
decreasing order of `downtimes'; MTTR mean time to repair or MCBF main cycle between failures. 
When trying to fix a specific fault, the plant
maintenance group 
look at the frequency and downtime histories of that particular fault, 
trying to ascertain whether
this fault has repeatedly caused problems in the past 
(and if it has, the engineers investigate
the maintenance records to find out
how the problem was fixed on those occasions and try out the same or improved solutions).
However, all these kinds of so-called single variable
analyses were only capable of capturing individual fault characteristics.
Fig.~\ref{fig:single-variable} shows a typical screen shot of the fault statistics
that the plant maintenance group can access when trying to fix some fault 
condition in the assembly plant.
However, the problem that the plant maintenance group face on a regular basis
can be much more complex. At times of machine faults, sometimes the last 
fault that brings the line to a halt is neither the
root-cause nor is it sufficiently indicative of it. Further,
any kind of summary of the machine faults-log
based on simple single variable statistics is limited in
its expressive power, and is unable to be of help to the plant maintenance group.
An important analysis tool in such situations would be one that can unearth
temporal correlations {\em between} faults. (Based on past experience,
it is well known that such correlations exist, although these are not immediately
apparent from the faults log). By looking at the logs, it should be
possible to ask questions like, is there a frequently occurring fault, say B, that
typically follows the fault, A, within a 5 minute interval? Are there any such significant
correlations among the faults being logged, that is indicative of 
where the actual root-cause of the current
fault may lie? For example, if most occurrences of A are followed by a B, 
or alternately most
occurrences of A are preceded by a B, then it is quite likely that one of them is a
root-cause of the other.

The main difficulty in estimating the relevant joint statistics from the data 
is that the total number of fault possibilities are often quite large and
it becomes computationally infeasible to systematically
estimate all possible correlations in the
data. This makes data mining an ideal tool for such analysis. 
Data mining algorithms are tailor-made
to efficiently estimate only the  strong correlations that stand out in the data 
(by saving computational effort that would otherwise have been wasted in 
 estimation of all the weaker correlations). In particular, since our
application demands the search for time-ordered correlations 
(that could indicate, e.g., whether A follows
B or B follows A) we resort to temporal data mining algorithms 
for analysis of the  engine assembly plant data.

\subsection{Temporal data mining}

Temporal data mining  \cite{RS99,LS05,moerchen07} is concerned with the exploration
of large sequential databases for hidden and unsuspected structures or
patterns that are (typically) previously unknown, but interesting
and useful to the data owner. Sequential databases are available
in several application domains ranging from stock market analysis
to bioinformatics \cite{EG01,TSD00,ALSS95,AS95,MTV97}.
Here our interest is in the machine faults log which is a time-ordered
sequence of faults that occurred in the line. 
By analyzing these large
volumes of data we can find useful temporal correlations among fault occurrences which 
can, in turn, be used for root cause diagnostics. 
%In this paper we describe how we can adopt some temporal
%data mining techniques \cite{srivats2005,srivats2007} in such a manufacturing domain. 

As stated earlier, a machine fault log is  a time-ordered sequence of faults
that occurred in a line, where each fault is recorded using a {\em code}
from an extensive (and often hierarchically arranged) alphabet. From a
data mining perspective, it is of interest to ask which combinations
of faults tend to occur frequently together with or without some
prescribed time constraints between occurrences of different faults. Such
analysis can be of great help in identifying root causes for recurrent
problems and hence is
useful for suggesting process improvements.  In this paper we show
that the framework of frequent episode discovery \cite{MTV97,LSU05,LSU07}
is well-suited for identifying such recurrent fault correlations in
the machine fault logs. We describe a temporal data mining system for
analyzing logs from engine plants using the frequent episodes framework.

The framework of frequent episodes in event streams provides a good
abstraction for mining useful temporal patterns from time-ordered data
\cite{MTV97}.  An episode is a short ordered sequence of {\em
events} where each event is tagged with an event-type from an appropriate
alphabet. In case of the machine faults data, the alphabet is the set
of codes required to uniquely locate and identify the fault in the
line. An episode is said to occur in a given data stream if the events
that constitute the episode appear in the data with the same order as
prescribed in the episode. The frequency of an episode is a measure of
how often the episode occurs in the data and episodes whose frequency
exceeds some threshold are declared as {\em frequent episodes}. We
consider the problem of discovering all frequent episodes in a given
sequence of faults logged.
 Frequent episodes are readily interpretable as significant fault
co-occurrences which are useful in the analysis of the fault logs. Using
these frequent episodes, it is possible to generate a list of online
alerts to help locate root causes for persistent problems. Since the
list of alerts is much smaller than the original machine fault logs, and
since it is also easier to read and interpret, it has been found to be
of great use to the plant engineers for fault diagnostics. The temporal
data mining techniques described in this paper have been incorporated
into a fault diagnosis system which is used on a regular basis on the
work floor of one of GM's engine plants (and is scheduled for adoption 
in several other plants).

The paper is organized as follows.  Sec.~\ref{sec:tdmf} presents details of
the data mining framework used for mining the fault logs. 
Sec.~\ref{sec-3} describes how to incorporate prior knowledge of the plant floor engineers 
into the application and how to structure the output of the data mining analysis 
in order to make it useful for the plant engineers. 
In Sec.~\ref{sec:r}, we discuss some sample results obtained to highlight
the effectiveness of temporal data mining in this application.

\section{Temporal data mining of machine fault logs}
\label{sec:tdmf}

%\subsection{Fault sequences from fault logs}
%\label{sec:fsffrl}

The machine fault logs used in this application were from three lines
in the engine assembly plant -- the engine block machining line, the
engine head machining line and the final engine assembly line. Based
on the physical layout of machines, it is known that there can be no
meaningful correlations among machine faults across these different
lines. Therefore, as a preprocessing step, the data is partitioned
into these three categories before analyzing it. Once this is done,
fault records within each category can be sorted based on the times of
occurrence to yield {\em fault sequences} that can each be separately
subjected to temporal data mining analysis.

%\subsection{Utility of temporal data mining}
%\label{sec:uotdm}

As mentioned earlier, 
%the main objective of applying temporal
%data mining methods on the fault sequences is to unearth temporal correlations
%hidden in them. 
using the frequent episodes framework it is possible to
discover all fault correlations that occur repeatedly in the data. Some
of these correlations may already be known to the plant engineers,
while some others may not. It is the unknown correlations that  plant
engineers are most interested in. For the data mining effort to be effective
in such an application, the episode discovery algorithms have to be properly
augmented with heuristic knowledge of plant engineers. This is needed to make
sure that the final output is useful and the plant engineers are not flooded 
with many irrelevant
correlations. 

In this section, we provide an overview of our temporal data mining method. Since the 
 algorithms are reported elsewhere \cite{LSU05,LSU07} we do not provide details of the 
algorithms here. In the next section we discuss how we incorporate some application specific 
knowledge and how we provide the final outputs of the data mining analysis.

%In the sections to follow we
%detail how fault sequences can be analyzed and plant engineers alerted
%so that some unexpected correlations that happen to be present in the
%data are brought to their notice at the earliest. This way, the number of
%(and also the durations of) stoppages in the lines can be significantly
%reduced, thereby enhancing the throughput in the assembly plant.

%Secs.~\ref{sec:eies}-\ref{sec:oudip} below detail how this
%can be done. Secs.~\ref{sec:mzaif}-\ref{sec:qcofc} describe how the final
%output fault correlation reports are prepared by filtering and classifying
%the set of frequent episodes discovered. Finally, in Sec.~\ref{sec:rgaa},
%we present the alert generation mechanism employed for online fault
%diagnosis in the engine assembly plants.

\subsection{Frequent Episodes in event streams}
\label{sec:eies}

%In our application, the fault sequences obtained in Sec.~\ref{sec:fsffrl} are analyzed
%using the frequent episodes discovery framework. In this section,
%we first briefly describe our framework of frequent episode discovery
%and give an overview of the data mining algorithms we use. Since these
%algorithms are reported elsewhere \cite{LSU05}, we do not provide full
%details here. After that we describe how the faults  data is processed
%and the final outputs are generated by the system.

The framework of frequent episodes is introduced in \cite{MTV97} as a generic 
method to discover certain temporal patterns in categorical time series data. 
The input to the frequent episode discovery algorithm 
\cite{MTV97} 
is an {\em event sequence}, $s = \langle (E_1,t_1),$
$(E_2,t_2) \ldots \rangle$, 
where each $(E_i,t_i)$ denotes an {\em event}
with {\em event type}, $E_i$, coming from some finite alphabet of event types and $t_i$
representing the time of occurrence of the event. For the engine
plant data, there are various ways to constitute the alphabet of event types. For
example, just the operation code in each fault record could be used
as the event type and an event sequence could be constituted by simply
considering the sequence of machines that reported faults. Alternately,
if we want to describe faults in greater detail, the subsystem and/or
fault code(s) may also be appended to the operation code to obtain the
event type corresponding to each fault record. The alphabet in such a
case would be the collection of all possible combinations of operation,
subsystem and fault codes. The temporal data mining toolbox that was
developed for this application provides
the user with both options.

An {\em episode}, $\alpha$, is an ordered\footnote{This corresponds to
the {\em serial} episode in the framework of \cite{MTV97}.  In general, 
an episode is essentially a collection of event types with a partial order over them. In
our application we are not interested in episodes with any other partial order 
among the nodes and hence we use the term episode to describe serial episodes.}
collection (or sequence)
of event types, and is denoted by $\alpha = (A_1 \rA \cdots A_N)$, where
each $A_i$ is an event type, arrows define the ordering among the event
types and $N$ denotes the size of $\alpha$. 
Further, any subsequence of
the episode $\alpha$, defines a {\em subepisode} of $\alpha$. 

%In order to allow for efficient
%frequent episode discovery, it is important that its frequency is always
%bounded above by the frequencies of its subepisodes. 

Episode $\alpha = (A_1 \rA \cdots A_N)$ is said to occur in a data sequence,
$s$, if the event types, $A_i$, that constitute it appear in $s$ in the
same order as in $\alpha$. We note here that the events of type $A_1,A_2$ etc. do not 
have to appear {\em consecutively} for the episode $\alpha$ to occur. There can be 
other events in between. 

For example, the following is
an event sequence containing ten events:
\begin{equation}
\langle(A,3),(D,4),(B,5),(C,9),(E,12),(A,14),(F,15),(B,18),(D,19),(C,27)\rangle
\label{eq:seqex}
\end{equation}
There are four occurrances of the episode $A\rightarrow B \rightarrow C$ in this 
data sequence. 

In the engine data, an episode is simply a collection of faults occurring
in a time-ordered fashion. Thus, the structure of episodes readily captures
temporal correlations among faults in a simple manner.

The objective of the data mining process here is to discover all {\em frequent episodes} 
where an episode is frequent if its {\em frequency} exceeds a threshold. The frequency 
of an episode is some measure of how often it occurs in the data sequence. 
There are many ways to define the episode {\em frequency} \cite{MTV97,laxman06}. 
In general, it is computationally inefficient to count all occurrances of an 
episode \cite{laxman06}. 
The motivation for defining different frequency measures is to be able to 
efficiently count the frequencies of a set of candidate episodes through a single pass 
over the data stream while ensuring that higher frequency would mean higher number 
of occurrances of an episode.   
In \cite{MTV97},
frequency is defined as the number of fixed-width sliding windows
over the data that contain at least one occurrence of the episode.  
In the example sequence given by (\ref{eq:seqex}), if we take a window width of eight, 
then the frequency of $A\rightarrow B\rightarrow C$ is two. (This is because the occurrance 
constituted by the events: $(A,3),(B,5),(C,9)$ is present in two windows, namely, 
$[2,\;9]$ and $[3,\; 10]$ while none of the other occurrances can fit in a window of 
width eight). In
\cite{laxman06}, the frequency is defined as the maximum number of {\em
non-overlapped} occurrences of the episode in the sequence, where, two
occurrences are said to be non-overlapped if no event associated with
one occurrence appears in between those of the other. In the example 
sequence given by (\ref{eq:seqex}), there are two non-overlapped occurrances 
of the episode $A \rightarrow B \rightarrow C$. (These two occurrances are constituted 
by the events: $(A,3),(B,5),(C,9)$ and $(A,14),(B,18),(C,27)$). The non-overlapped
occurrences-based frequency is  computationally much more efficient
than the windows-based count \cite{LSU05,LSU07a}. It is also theoretically more elegant. 
It allows for a formal connection between episode discovery and learning
of stochastic generative models (for the data source) in terms of some
specialized Hidden Markov Models (HMMs) \cite{LSU05} and this, in turn, allows 
one to easily assess the statistical significance of the discovered frequent 
episodes.  Also, in our application of analyzing fault sequences, if episodes are 
to capture some underlying causative temporal correlations, counting only non-overlapped 
occurrances is intuitively appealing. Hence, in our 
 application, we adopt this non-overlapped occurrences-based
count as the frequency definition for episodes. An  episode with {\em
high} frequency essentially indicates a {\em strong} correlation among
an ordered sequence of faults in that it happen repeatedly and hence may be a 
useful indicator of some underlying fault condition. Often, we may prescribe an 
additional constraint, called expiry time, under which we count an occurrance 
only if the time span of the occurrance is less than some prespecified threshold. 
In the earlier example, if we have an expiry time of 8 time units, then there is 
only one non-overlapped occurrance of the episode $A\rightarrow B\rightarrow C$. 
We note here that the window width in the windows-based frequency \cite{MTV97} 
can be thought of as an expiry time constraint. However, if the actual span of the 
occurrance is much less than the window width then an episode with only one occurrance 
can still have a large frequency because it stays in many consecutive windows. 
This problem is not there with the non-overlapped occurrances based frequency. 
Moreover, it is computationally more efficient to count non-overlapped occurrences and we
use this frequency definition in our application.

% When  frequency of an episode exceeds some
%threshold it is declared a {\em frequent episode}.

%In our data mining system implemented in the plant, we actually use a
%generalized episodes framework \cite{LSU02}. Here, each event in the
%data stream is described by an event type along with both a start time
%and an end time for the event. The generalized episodes prescribe not
%only the event types but also duration constraints for the events that
%constitute an occurrence of the episode. However, the notation needed
%to describe these is somewhat more complicated. Since, in this paper,
%our interest is in describing an application of the temporal data mining
%method, we stick to the simpler version of episodes as described above.

\subsection{Algorithms for frequent episode discovery}
\label{sec:affed}

The discovery of all frequent episodes in an event stream can be
efficiently carried out using a level-wise Apriori-style procedure that 
is popular in most of the frequent-pattern-discovery methods in data mining. 
Such a method is feasible if the frequency measure used is such that the 
frequency of an episode is less than or equal to that of each of its subepisodes. 
(Both the frequency measures mentioned above have this property). This gives rise to 
the key observation: the necessary (though not sufficient) condition for an $N$-node 
episode to be frequent is that each of its $(N-1)$-node subepisodes are frequent. This 
is very effective in controlling the combinatorial explosion in generating a set of 
$N$-node candidate episodes for frequency counting as explained below. 
The method of frequent episode discovery consists of performing  
 the two steps of {\em candidate generation} and {\em frequency
counting}, repeatedly,  once for every successively larger size of
episodes.  First, all {\em frequent} episodes of size $1$ are found by
building a simple histogram for the various faults that occurred in the
data. These are then combined to obtain {\em candidate} episodes of size
$2$ using a candidate generation procedure \cite{MTV97}.  
The next step involves
frequency counting of the candidates just generated and this is done 
through one pass over the data stream 
using the finite state automata based algorithm as 
described in \cite{LSU05,LSU07,LSU07a}.  Once
frequent episodes of size $2$ are thus obtained, they are used to construct
candidate episodes of size $3$ and by one more pass over the data we count the frequencies 
of all the candidates to obtain the frequent 3-node episodes and so on. 
This process is repeated till
eventually, episodes of all required sizes are discovered.

The objective of candidate generation algorithm is to present to the 
frequency counting step, as few candidates as possible without missing 
any episode that would be frequent. 
As explained earlier, an episode can be frequent  
 only if all its subepisodes were earlier found
frequent in the previous level of the algorithm. This is what is 	
exploited in the candidate generation step for constructing $(N+1)$-node 
candidates from $N$-node frequent episodes. We build the candidates by 
taking all possible pairs of $N$-node frequent episodes that have $N-1$ nodes common and 
combine each such pair to yield $(N+1)$-node candidates. Such a candidate generation 
procedure can control the combinatorial explosion because, as the size of episodes 
grows, the number of frequent episodes falls rapidly (if we choose our frequecy 
threshold well).  
In the frequency counting algorithms, occurrance 
of any episode in the data is recognized by having finite state automata for 
that episode. For example, the automaton associated with the episode 
$A\rightarrow B \rightarrow C$ would first wait for an event of type $A$ so as to transit 
into its first state and then wait for an event of type $B$ and so on. When 
the automaton transits into its final state, one occurrance would be recognized. 
Since there are many candidate episodes and since we may have to track multiple 
potential occurrances of each episode, there would be many such active automata 
at any time. The algorithm consists of going through the event sequence and 
for each event, efficiently managing all the needed state transitions of these 
automata. The details of the algorithms are available in \cite{LSU05,LSU07}.

\subsection{Handling events with non-zero durations}

In the formalism described so far, it is implicitly assumed that events are 
instantaneous. That is why, in the event sequence, each event is associated with only 
one number denoting its time of occurrance. However, in our application, different 
faults persist for different durations. The faults sequence data contains two time 
stamps for each record, namely, the start time and the resolved time for each fault. 
The durations for which different faults persist is, in general, important in unearthing 
useful temporal correlations. Hence we need to extend the formalism to the case where 
different events persist for different durations of time. Such an extension has been 
developed \cite{LSU07} and this is the framework that is actually used. (This 
extended formalism is, in fact, motivated by the application described here).

In the extended framework, the event sequence is of the form 
 $s = \langle (E_1,t_1,\tau_1),$ $(E_2,t_2,\tau_2) \ldots \rangle$, where $E_i$ is the 
event type as earlier and $t_i$ and $\tau_i$ denote, respectively, the start and 
end times of $i^{\mbox{th}}$ event. We will call $(\tau_i - t_i)$ as the dwelling 
time of $i^{\mbox{th}}$ event. The episodes in the new framework, called generalized 
episodes, contain, in addition to an ordered sequence of event types, a set of time 
intervals that prescribe the allowed dwelling times for events that constitute an 
occurrance of the episode.  For example, a two node episode here could be represented 
as $A(I_1) \rightarrow B(I_2)$. For an ocurrance of such an episode in the data 
stream we need an event of type $A$ whose dwelling time is in the interval $I_1$ which is 
followed some time later by an event of type $B$ whose dwelling time is in the 
interval $I_2$. (In general, it is possible to associate a finite union of intervals 
with each node in the episode). The time intervals that we can associate with any node 
in an episode come from a finite collection of disjoint time intervals provided by the 
user. This set of intervals essentially prescribes the different time durations for 
events that are sought to be distinguished and can be used to analyze the data stream in
different time granularities. 

For the generalized episode mining also we use the same frequency, namely the number 
of non-overlapping occurrances of the episode. The two step procedure of 
candidate generation and frequency counting is the method used for discovering 
frequent generalized episodes also. 
However,both candidate generation as well as frequency 
counting become  more complicated because of the need to handle time durations 
of events. The details of the algorithms can be found in \cite{LSU07}.  

\subsection{Significance of episodes and frequency threshold}
\label{sec:soe}

As mentioned earlier, frequent episodes are those whose frequency is above 
a threshold. Thus, the frequency threshold is a critical parameter for the 
frequent episode discovery algorithm. In this subsection we mention some 
theoretical results based on which it is possible to automatically arrive at a 
reasonable frequency threshold.  

Earlier we noted that episodes with higher
frequencies are likely to represent more important correlations among a set of 
repeatedly occurring faults and hence are more
useful to the plant engineers during fault diagnosis. This aspect has
been actually formalized in a statistical sense in \cite{LSU05}. This
is done by defining a special class of Hidden Markov Models (HMMs)
called Episode Generating HMMs (EGHs) and defining an association from
episodes to EGHs. It was proved that under this association, the more
frequent episode is associated with the EGH that has higher likelihood
of generating the given event sequence. (Here, frequency is the number of 
non-verlapped occurrances of the episode). This theoretical connection
facilitates a test of statistical significance for frequent episodes discovered. 
 This test is a simple one
that requires nothing more than the output of the frequency counting
algorithm to determine significance of an episode. More specifically,
for a given probability of error, an episode must have a particular
{\em minimum} frequency in the event stream for it to be regarded as
significant. This minimum frequency needed depends only on the number
of nodes in the episode, length of the data stream, the size of the
alphabet for describing event types and the allowed probability of
error. For error probability less than $0.5$, the minimum frequency
needed does not vary much with the error probability and it is close
to $(\frac{T}{M N})$, where $T$ is the length of event stream, $N$
is the size of the episode and $M$ is the size of the alphabet. This
is therefore a good initial choice for frequency threshold in frequent
episode discovery and we always use this for preliminary analysis of
data. (See \cite{LSU05} for details of this analysis). 
In fact, this threshold is often found to be very good even for
final analysis, although the user always has the option to set a higher
threshold for frequency whenever necessary. The theoretical analysis 
for arriving at this frequency threshold (as presented in \cite{LSU05}) is 
valid only for the case of instantaneous events. However, even in case of 
events with time durations (and hence for generalized episode discovery), the 
same threshold is found to be very effective in our application. 

\subsection{Other user-defined input parameters}
\label{sec:oudip}

In general, our frequent episode discovery algorithms (for 
instantaneous events) do not require any
user-defined input parameters. While the candidate generation algorithm
of \cite{MTV97} only requires the set of frequent episodes from the
previous level as input, the frequency counting algorithm of \cite{LSU05}
requires the event stream, the current set of candidates and a frequency
threshold as input.  As was just described in Sec.~\ref{sec:soe},
the frequency threshold can be set automatically.

As stated earlier, in our application the durations of events are important 
and hence we actually use the generalized episode framework. Hence, one 
input needed from the user is the set of time durations that are sought 
to be distinguished. In addition, it is often found useful to impose one 
more temporal constraint on the episode discovery which we call expiry time 
constraint. These are explained in this subsection. 

%However, when analyzing the GM engine plant data it has been found
%useful to add some extra constraints on the kind of episodes being
%counted. These constraints, which we refer to as the {\em expiry time}
%and {\em duration interval} constraints, are explained below:

\begin{itemize}

\item Expiry time: As per our definition of episode occurrence, even events 
separated by arbitrarily large time interval can still constitute occurrence of 
an episode. However, faults (which are the events for us) that occur far from 
each other are unlikely to have any causative relationship and hence we do not want to 
count occurrences of episodes which are constituted by such events. 
To ensure that faults  widely separated in 
times are not counted as an occurrence of an episode, it
is possible to prescribe an {\em expiry time} for episodes. This is
a user-defined parameter which bounds the time difference between the
first and last events within a single occurrence of the episode. Thus, now the 
frequency of an episode would be the maximum number of non-verlapped occurrances 
of an episode such that each occurrence satisfies the expiry time constraint. This
extra constraint is easily incorporated into the frequency counting
procedure as is indicated in \cite{LSU05}.

\item The set of possible durations: For mining engine assembly data,
based on inputs from plant engineers, we bucket the time durations of events 
into the following four intervals:
$[1\sm 120]$, $[121\sm 600]$, $[601\sm 1800]$ and $[>1800]$. The s 
behind these time durations are the dynamics of fault recovery. The duration   
of the event depends on the fault types and the tactical decision by the 
maintenance group. Further, plant engineers are not
interested in long events of one particular type, or short events 
of another type, etc. The
generalized episodes discovery algorithms of \cite{LSU07} are capable of 
handling all such special
cases within the same unified framework, and these algorithms
are incorporated into the temporal data mining toolbox.

%In addition to considering the above set of time duration possibilities, there 
%are a few other preprocessing steps needed. 

%In fact, durations of events can play a greater role in
%the temporal pattern discovery process. It is possible
%that short events of type $B$ may follow long events of type $A$, etc. 
%In our engine assembly plant application,
%time durations of events are very important for plant
%engineers and these kinds
%of duration-dependent fault patterns have been found useful when
%carrying out fault diagnosis. 
%The framework of frequent episodes can be generalized \cite{LSU02}
%to allow for episodes to explicitly describe event sequences with
%multiple time duration constraints. 
%In this framework, episodes are patterns that look like
%$A(601\sm1800)\rA B(1\sm120)$, which describes a correlation between events of 
%type $A$ whose
%time durations lie in the interval $[601\sm1800]$, with events of type $B$ 
%whose durations lie in
%the interval $[1\sm120]$. The generalized episodes framework requires 
%additional input from the user
%in the form of a set of time intervals that can be used to describe such patterns. 

\end{itemize}

\section{Incorporating plant floor domain knowledge}
\label{sec-3}

In the previous section, we described how the framework of frequent episode
discovery allows for a theoretically elegant and computational efficient
temporal analysis of machine fault logs in automative engine assembly
plants. However, it is important to note that it is a model-free technique
and as is the case with any such method, the technique will be really
useful, only after considerable effort is put in to incorporate
explicit domain-specific knowledge into the data mining system. We perform
this through pre-filtering of the data and post-filtering of the results from
frequent episode discovery. In this section, we describe some of the details of
how such information is incorporated into our application. .

\subsection{Pre-filtering the input data}
\label{sec:prefilter}
For a data mining technique to be effective in any application, it is always necessary 
to `clean up' the data and filter out the noise. This will help in ensuring that the output of the data mining 
analysis is useful and is not cluttered with too much of irrelevant information. 
%Here 
%we explain some of these operations incorporated into our application. 

The machine faults logged 
sometimes contain `spurious' records in the form of some {\em zero
duration} faults. These records occurred due to communication problem between 
the machine controller and the data network data collection system or due to 
programming error in the controller. 
These records need to be removed from the data before
analysis. Also, plant floor engineers are not interested in unearthing
correlations involving faults with large durations (i.e.~time difference
between resolved time and occurrence time of the fault is large). Both these
duration constraints can be handled by requiring that the duration of
any fault considered for analysis lies within some user-defined time
interval. For example, if the user specifies this duration constraint
by the time interval $[1\sm 1800]$, this would mean that only non-zero
duration faults with duration within 1800 seconds are considered and all
other fault records are filtered out of the episode discovery process. Events 
that create faults of very long durations are typically not correlated with any other faults. They
are random events depending on variables such as buffer status in the machining lines or unscheduled 
breaks in the assembly lines.

%It is fairly straightforward to implement this by simply checking for this
%constraint before allowing any fault in the faults log to be considered as
%an event in the event stream. 

Another important prefiltering operation is to determine 
%This filter is applied to the data that is input to the frequent episode discovery algorithms.
%The pre-filter first determines 
the granularity of the alphabet (or codes) used to
describe the data. As was mentioned earlier in Sec.~\ref{sec:tdmf}, 
the event sequence input to
the frequent episode discovery can be constituted in many ways. 
We could either use or ignore the
fault codes in the fault logs, and correspondingly, the data mining 
analysis would operate at either
the faults level or station (or machine) level. In addition, 
there are many logical ways to group the machines in
the assembly plant - line-wise, zone-wise, based on whether the machine is manual 
or automatic, etc.
One could either choose to analyze all of the data that is logged or simply choose data
corresponding to one or more logical groups. If the user is looking 
for patterns specifically 
within one such logical group, then applying this restriction as a 
prefilter can significantly help
speed-up the process of frequent episode discovery due to a reduction in 
length of the data stream.

Another important prefilter that is applied on the data sequence is to 
remove some specific fault
codes that the plant engineers know carry no serious information. 
For example, some machines may be
programmed to go down when some other fault occurs. These fault occurrences 
would be part of some
kind of plant logic or machine logic and it is useless to look for 
patterns involving one or more of
these. These cases include machine faults such as E-Stop faults, I/O faults and communication faults. So we remove all such machine and/or fault codes 
before starting any data mining analysis.

\subsection{Domain specific heuristics: Multiple machine and individual machine faults}
\label{sec:mzaif}

Most data mining methods that aim at discovering frequent patterns usually
come up with a large number of frequent patterns. This is particularly
true in large application domains such as the one we consider here.
Hence, to make the method effective and useful to the plant engineer,
we have to use application specific heuristics to focus attention on
patterns that are likely to be of interest. These heuristics are based
on the knowledge that plant engineers have about the assembly
process. 
%In this subsection we describe one such method of restricting
%the kind of episodes that we look for in the data.

The frequent episode discovery process throws up a wide variety of
fault correlations as output.  Of these, from the point of view of
fault diagnosis and root cause analysis, there are two kinds of fault
correlations (or episodes) that are particularly interesting. We refer
to them as {\em multiple machine faults} and {\em individual machine faults}. A
multiple machines fault correlation refers to one involving at least two
different machines. Multiple machine fault correlations can 
in-turn be of two types, namely, those
that involve multiple machines along with their zone controller, 
and those that involve neighboring
machines in general. By contrast, an individual machine fault correlation
refers to a sequence of faults reported from the same machine.
Our episode discovery method is programmed to look for only those episodes that 
satisfy these structural restrictions. 

An important aspect of this categorization of fault correlations is that
plant floor engineers require to use different parameter values when 
looking for these different
types of fault correlations. For example, the expiry time for individual 
faults is computed as the
difference between the end-time of the first event and the start-time 
of the last event in the
episode's occurrence; whereas, for machine faults, the expiry time is 
taken as the difference
between the start-times of the first and last events in the episode's 
occurrence. The reason for
this is as follows. When considering individual faults, 
we are looking at faults occurring within the same
machine and the same machine cannot simultaneously be in two states. 
So it is reasonable to consider
the time duration between the end of the first and the 
beginning of the last event in the
occurrence. In case of multiple machine faults, the events in the 
occurrence can very well overlap
and hence the difference between start times is a more useful indicator. 
Our application incorporates  
all such special flexibilities. 
%this makes it difficult to interpret an expiry time defined using both start and end times.
%Since we order events in the data based on start times, for machine faults we compute expiry times
%also based purely on the start times.

From the point of view of the user, the utility of restricting the output to only 
episodes with these special kinds
of structures (namely machine faults and individual faults) is that
the number of frequent episodes output is significantly
reduced thereby enhancing the readability of the final fault correlation
reports. 
These kinds of fault correlations, in  a sense, 
lead to `actionable' output from the data
mining algorithm. When plant floor engineers look at 
correlations other than these, they find it
difficult to interpret them and so they cannot 
find use for them during fault diagnosis.

\subsection{Structure of the Output: Rule generation and alerts}
\label{sec:rgaa}

\begin{figure}[t]
\centering

\includegraphics[height=5in]{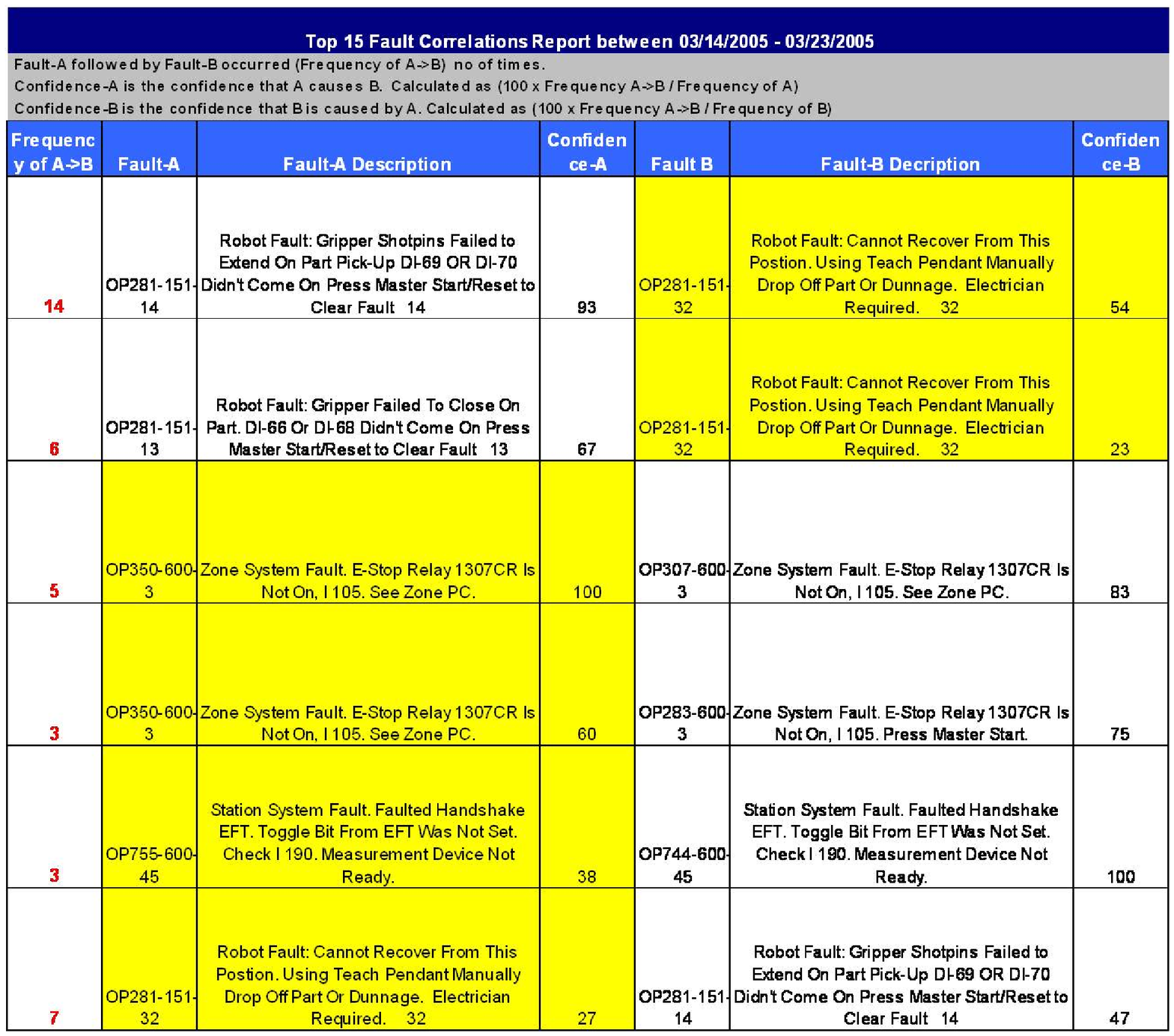}
\caption{A typical output report based on the 2-node episodes discovered by the tool. Frequency of the 2-node
episode is reported in Column~1. Each pattern (or 2-node episode) in the list is of the
form, $(A \rA B)$. The error code corresponding to event $A$ (i.e.~the first node of the 2-node episode)
is given in Column~2, and a description of the corresponding error is given in Column~3. 
Columns 5 and 6 describe the same things for event $B$ (i.e.~the
second node of the 2-node episode). Confidence of the rule ``$A$ causes $B$'',
$(100\times\mathrm{Frequency\ of\ } A\rA B)/(\mathrm{Frequency\ of\ }A)$, is reported in
Column~4. Similarly, confidence of the rule ``$B$ causes $A$'', 
$(100\times\mathrm{Frequency\ of\ } A\rA B)/(\mathrm{Frequency\ of\ }B)$, is reported in Column~7.}

\label{fig:typical-tdminer-output}
\end{figure}

An important aspect of any data mining application is the issue 
of what should be the output of 
the analysis. The frequent episode discovery algorithm 
outputs a list of frequent episodes 
discovered in the data stream. Hence, one possibility is to present this list 
in a suitably sorted order. This is one of the modes in which our application can be run. 
Everyday, the frequent episode discovery is run on a window of 
past data (e.g., the immediate 
past one week) and the list of frequent episodes (which denote some temporal correlations 
detected), are output to the user. Fig.~\ref{fig:typical-tdminer-output} shows a typical output from our 
application listing some frequent 2-node episodes. Such a list of frequent episodes is 
found to be a useful routine report for the plant engineers. 
This output also shows which of 
the frequent episodes involve faults that contributed most for the downtime during this 
time period. The frequent episodes in this output are sorted not only 
by their frequency but also based on what we call as scores of the episodes. The score of frequent episode is 
a new measure of the possibile utility of a discovered episode and it is explained below.

In general, the frequent pattern discovery in data mining is aimed
at generating useful rules to capture regularities in the data.
The frequent episodes discovered can be used to construct rules like
``$\beta$ implies $\alpha$'' where $\beta$ is a subepisode of $\alpha$
(and both are frequent) \cite{MTV97}.  The frequency threshold used
for episode discovery is basically a minimum support requirement for
these rules. The {\em rule confidence}, as always, is the fraction of
frequency of $\beta$ to that of $\alpha$. We use such rule formalism
to come up with two more heuristic figures of merit (in addition to the
frequency) for each of the frequent episodes.  For this we consider,
for each frequent episode, $\alpha$, rules of the form ``$\alpha^{(i)}$
implies $\alpha$'' where $\alpha^{(i)}$ denotes the subepisode of $\alpha$
obtained by dropping its $i^\mathrm{th}$ node. We define the {\em best confidence score}
of a frequent $N$-node episode, $\alpha$, as the maximum confidence among
rules of the form ``$\alpha^{(i)}$ implies $\alpha$'' for  all $i = 1,
\ldots, N$. An episode's best confidence score, being simply the confidence of the best
rule that it appears in, is thus a measure of its inferential power.
Similarly, we define the {\em worst confidence score} of a frequent $N$-node
episode, $\alpha$, as the minimum confidence among rules of the form 
``$\alpha^{(i)}$ implies $\alpha$'' for  all $i = 1,
\ldots, N$. This figure of merit, in a sense, measures the 
strength of the weakest inference that
can be made using the given episode.
Recall that, when analyzing the GM data, we already use one threshold for the frequency
(which is basically an input parameter to the frequent episodes 
discovery process). In addition,
now, we specify two more thresholds for the best and worst confidence 
scores respectively.
Only if the frequency exceeds the frequency threshold and both the 
best and worst confidence scores,
exceed their own corresponding thresholds, will an episode be 
finally presented as an output 
to the user.

As stated earlier, one of the modes in which our system runs 
is to discover and present to
the user a set of frequent episodes (with their frequencies and scores)
over a data slice. Another useful 
and novel mode of employing frequent episode discovery in this application 
is what we call an alert generation system. This system can be deployed as a software
agent integrated in a computerized maintenance management system such as 
MAXIMO. This integration can be significant to react to the identified 
faults correlations and reduce the risk of losing throughput.

Our alert generation system, based on frequent episode discovery, indicates, 
on a daily basis, 
%application of the frequent episode framework in GM's engine assembly
%plants is an alert generation system that indicates, on a daily basis,
which episodes seem to be dominating the data in the recent past. This
alert generation system is based on a sliding window analysis of the data.
Every day, the frequent episode discovery algorithms are run for faults
logged in some fixed-length history (say, data corresponding to the
immediate past one week). Since this is done on a daily basis, for each
day we have a set of frequent episodes and their scores.  Based on the
frequencies and scores of the episodes discovered over a sequence of
days, we use the following heuristic to generate alerts:  ``Whenever an
episode exhibits a non-decreasing frequency trend for some period (of say
4 days) with its frequency and score above some user-defined thresholds,
generate an alert on that episode.'' Thus, each day the system provides
the plant engineers with a list of alerts for the day. This list has been
found very useful for online fault analysis and root cause diagnosis.

\section{Results}
\label{sec:r}

As indicated earlier, our method  of frequent episode discovery along
with the heuristics (based on knowledge of plant engineers) to prune
the set of frequent episodes, resulted in a useful tool whereby a small
enough set of episodes are presented to the user.  Such a system is
currently being used in one of the
plants and the outputs provided are found to be useful to the user. In
this section we provide some discussion on assessing the usefulness of
this system along with a couple of examples.

\subsection{Qualitative assessment of fault correlations discovered}
\label{sec:qcofc}

The system was first tested extensively on historical data before it was
actually deployed. Here we took data corresponding to a few months
and obtained the set of frequent episodes, which were then assessed for
``interestingness'' by the plant engineers.  The fault correlations
were qualitatively analyzed by using the engineers' experience and
prior knowledge about the manufacturing process. Three categories were
identified and each fault correlation in the output was classified into
one of the following:

\begin{itemize}

\item {\em Well-known episodes:} These correspond to correlations that
are very well-known and that routinely occur in the plant. Some of these episodes
may occur because, e.~g., the controller is programmed to bring down
some stations when a particular machine fails. These episodes carry
no new information and hence are not particularly useful to the plant
engineers.  However, it is seen that our algorithms regularly throw
up such episodes, which shows that the frequent episodes framework is
effective in identifying correlations that are known to exist in the
data. This resulted in building confidence of the plant engineers in the
capabilities of the system.  In the final system that we implemented,
a post-processing step flags these episodes as ``well-known'' and removes
them from the eventual list of episodes presented to the user.

\item {\em Expected episodes:} These correspond to correlations that
have been observed and learnt by the plant engineers through their past
experience in the assembly plant.  These episodes are useful during online
fault diagnosis but are already part of the troubleshooting heuristic
knowledge in the assembly plant. Our system has been throwing up many
such episodes also which is indicative of the framework's capability as
a fault diagnosis tool.

\item {\em Unexpected episodes:} These are correlations that past
experience in the assembly plant cannot immediately explain, and for
this very reason, are of greatest interest from a data mining view
point. The real utility of the frequent episode framework lies in
discovering unexpected episodes early, so that, the root cause of a
recurring problem can be quickly located and solved, thereby facilitating
better throughput in the assembly plant.  The algorithm was assessed
to be useful because it discovered a few such unexpected episodes in
historic data. These were unexpected in the sense that at the time
these occurred, the problem could not be immediately resolved. However,
subsequent troubleshooting has unearthed the root cause and thus, when
we ran our algorithms on the historic data, these episodes could be
recognized as very interesting correlations discovered by the algorithm.
We describe,  later in this section, some examples of useful unexpected
correlations discovered by the algorithm.

\end{itemize}

Through this process of assessment of frequent episodes output by the
algorithm we were able to conclude that the algorithm is a useful tool
for fault analysis and root cause diagnosis. It is noted here that
this classification of frequent episodes into well-known, expected and
unexpected episodes is a continuous ongoing process. With time and with
repeated indications of certain causative relationships among faults,
it is possible that episodes that were once regarded as unexpected can
now be classified as expected episodes. Such a situation may arise,
for example, when a new machine is installed in the plant, characteristics
of which are as yet unknown to the plant engineers. In the next section
we give an example of one such instance, when our algorithms unearthed
a correlation involving a new machine and which the plant engineers
considered as a very useful discovery.

\subsection{Some example episodes discovered}
\label{sec:seed}

As was mentioned earlier, the temporal data mining system described in
this paper is currently being used as a fault diagnosis tool on the
work floor of one GM's engine assembly plants. In this section,
we present a couple of non-trivial (unexpected) fault correlations that
were discovered in the data -- the first is a multiple machines fault
correlation unearthed during analysis of some historic data, and the
second is an individual faults correlation (cf.~\ref{sec:mzaif}) that
was found after the system was deployed online for fault diagnosis.

\begin{figure}[t]
\centering
\includegraphics[width=4in]{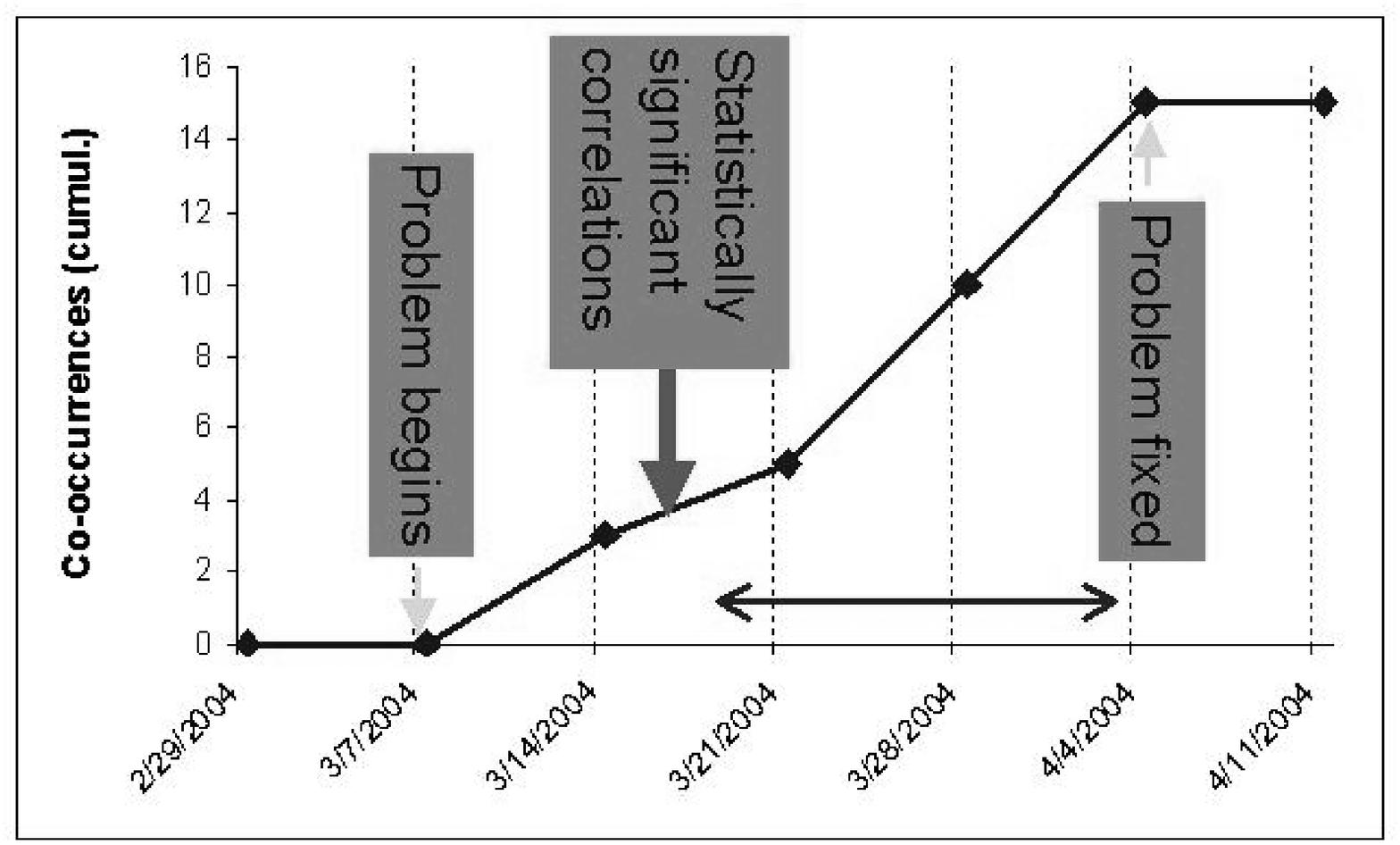}
\caption{Example of an interesting unexpected fault correlation discovered
in the data during historic analysis of machine fault logs from the period
of March-April 2004.}
\label{fig:co}
\end{figure}

When evaluating the utility of our temporal data mining techniques
for analysis of the engine plant data, the algorithms were run on some
historic machine fault logs from the period of January-April 2004. Here
is an example of a fault correlation in this data slice which could
have saved the plant engineers more than two weeks of troubleshooting.
Some time during the second week of March 2004, the plant started facing
throughput problems due to repeated failure in a particular station
(referred to as Station $A$ in this discussion). The root cause for this
problem was actually elsewhere in Station $B$ and this was eventually
identified only around the $4^\mathrm{th}$ of April 2004. An
analysis of the machine faults log (using our frequent episode discovery 
technique) showed that the episode $(B \rA A)$ was one of the 
frequent episodes during this period. 
%For each day,
%the frequent episode discovery algorithms were run on data starting
The analysis is pictorially depicted in
Fig.~\ref{fig:co} starting with $2^\mathrm{nd}$ March 2004.  
 The graph plots the cumulative frequency of episode
$(B \rA A)$ against the date of analysis. In the third week of March
2004, frequency of episode $(B \rA A)$ exceeds the frequency threshold,
$\frac{T}{MN}$, and hence is regarded by our algorithm as statistically
significant (cf.~Sec.~\ref{sec:soe}). Soon after, an alert was generated 
for this episode because it exhibited an increasing trend of frequency and the 
frequency and score requirements were met. Thus, if our temporal data
mining based fault analysis system had been used in the plant during
March-April 2004, the plant engineers could have been alerted about the
rising significance of this multiple machine faults correlation, about
two and a half weeks before it was eventually identified by them as the
root cause correlation. Based on the effective down-time of Station $A$
due to Station $B$, it is estimated that an additional 55 engines could
have been built during this time by avoiding these faults. This is an
example of one of the fault correlations whose discovery in the historic
data was influential in the final decision by the plant to adopt the
technique as a routine tool for fault diagnosis on the factory floor.

Next we present an example of an interesting individual faults correlation that was
discovered after our system was deployed for online use. 
 During March 2005, the plant was having problems because of
a robot (referred to here as Station $C$) going into a fault mode (with
fault code denoted by $X$ here) from which recovery was not possible
without manual intervention. The fault record `$C \_ X$' was being
repeatedly logged, but the engineers were unable to identify the root
cause because this robot, $C$, was newly installed in the plant and so
there was very little experience available about its failure patterns to
fall back upon. The temporal data mining algorithms, however, identified
the root cause of the problem through a significant frequent episode,
$(C \_ Y \rA C \_ X)$, in the data. Once the problem $Y$ was fixed in
the Station $C$, the fault $X$ stopped occurring and smooth running
of the line was resumed.

\section{Conclusions}

In this paper we have presented an application of temporal data mining
in the manufacturing domain.  Time-stamped faults logged in an engine
assembly plant are mined using the frequent episodes discovery
framework. The faults logged are first partitioned based on the physical
layout of machines in the  plant, to obtain fault sequences that can
be subjected to temporal data mining analysis. The frequent episode
discovery algorithms of \cite{MTV97,LSU05} are run on these sequences
and the {\em significant} frequent episodes (or fault correlations)
reported to the plant engineers to assist them in fault diagnosis. The
engineers are particularly interested in fault correlations with certain
specific structures and these heuristics are used to filter the frequent
episodes output by the system.

Our algorithms have been incorporated into a fault diagnosis toolbox in
one of GM's engine assembly plants. The performance of these algorithms
was first assessed on some historic data and was considered
useful by the plant engineers as an aid for routine troubleshooting
on the work floor. We have reported in this paper, one example of a
pattern discovered during historic data analysis. 
% which influenced the
%decision to use this technique as a fault diagnosis tool on  a regular
%basis. 
An interesting aspect of our data mining framework is that
there is no need for any detailed modeling of the underlying manufacturing
process. Apart from some simple heuristics based on the plant engineers'
experience, we do not use any other knowledge of the data generation
process. This advantage is highlighted in the second example result
reported in Sec.~\ref{sec:seed}, where the root cause of a problem was
in a new machine that was installed in the plant whose characteristics
was not yet known to the plant engineers. This result, we believe, is in
the true spirit of one the goals of data mining, namely, to unearth from
the data, useful (non-trivial) information that was previously unknown
to the data owner. 

To our knowledge, this is the first instance of the application of 
temporal datamining in the manufacturing domain. Due to the complex 
interactions among the components in any modern manufacturing system, such 
data mining analyses should prove to be useful in many other settings as well. 
We hope that the description of our application as reported here would help 
other practitioners in engineering such applications.

%\bibliography{srivats}

\begin{thebibliography}{10}

\bibitem{RS99}
J.~F. Roddick and M.~Spiliopoulou, ``A bibliography of temporal, spatial and
  spatio-temporal data mining research,'' {\em ACM SIGKDD Explorations
  Newsletter}, vol.~1, pp.~34--38, June 1999.

\bibitem{LS05}
S.~Laxman and P.~S. Sastry, ``A survey of temporal data mining,'' {\em
  $S\bar{A}DH\bar{A}N\bar{A}$, Academy Proceedings in Engineering Sciences},
  vol.~31, no.~2, pp.~173--198, 2005.

\bibitem{moerchen07}
F.~Moerchen, ``Unsupervised pattern mining from symbolic temporal data,'' {\em
  ACM SIGKDD Explorations}, vol.~9, pp.~41--55, June 2007.

\bibitem{EG01}
W.~J. Ewens and G.~R. Grant, {\em Statistical methods in Bioinformatics: An
  introduction}.
\newblock Springer-Verlag, New York, NY, 2001.

\bibitem{TSD00}
P.~Tino, C.~Schittenkopf, and G.~Dorffner, ``Temporal pattern recognition in
  noisy non-stationary time series based on quantization into symbolic streams:
  Lessons learned from financial volatility trading
  (url:citeseer.nj.nec.com/tino00temporal.html),'' 2000.

\bibitem{ALSS95}
R.~Agrawal, K.~I. Lin, H.~S. Sawhney, and K.~Shim, ``Fast similarity search in
  the presence of noise, scaling and translation in time series databases,'' in
  {\em Proceedings of the 21st International Conference on Very Large Data
  Bases (VLDB95)}, (Zurich, Switzerland), pp.~490--501, Sept 11--15 1995.

\bibitem{AS95}
R.~Agrawal and R.~Srikant, ``Mining sequential patterns,'' in {\em Proceedings
  of the 11th International Conference on Data Engineering}, (Taipei, Taiwan),
  IEEE Computer Society, Washington, DC, USA, Mar. 1995.

\bibitem{MTV97}
H.~Mannila, H.~Toivonen, and A.~I. Verkamo, ``Discovery of frequent episodes in
  event sequences,'' {\em Data Mining and Knowledge Discovery}, vol.~1, no.~3,
  pp.~259--289, 1997.

\bibitem{LSU05}
S.~Laxman, P.~S. Sastry, and K.~P. Unnikrishnan, ``Discovering frequent
  episodes and learning {H}idden {M}arkov {M}odels: {A} formal connection,''
  {\em IEEE Transactions on Knowledge and Data Engineering}, vol.~17,
  pp.~1505--1517, Nov. 2005.

\bibitem{LSU07}
S.~Laxman, P.~S. Sastry, and K.~P. Unnikrishnan, ``Discovering frequent
  generalized episodes when events persist for different durations,'' {\em IEEE
  Transactions on Knowledge and Data Engineering}, vol.~19, Sept. 2007.

\bibitem{laxman06}
S.~Laxman, {\em Discovering frequent episodes: {F}ast algorithms, connections
  with {HMMs} and generalizations}.
\newblock PhD thesis, Dept. of Electrical Engineering, Indian Institute of
  Science, Bangalore, India, Mar. 2006.

\bibitem{LSU07a}
S.~Laxman, P.~S. Sastry, and K.~P. Unnikrishnan, ``A fast algorithm for finding
  frequent episodes in event streams,'' in {\em Proceedings of the 13th
  International Conference on Knowledge Discovery and Data Mining (KDD'07), San
  Jose}, Aug. 2007.

\end{thebibliography}
%\bibliographystyle{ieeetr}

\end{document}